\pdfoutput=1

\documentclass[11pt]{article}

\usepackage[]{EMNLP2023}

\usepackage{times}
\usepackage{latexsym}

\usepackage[T1]{fontenc}

\usepackage[utf8]{inputenc}

\usepackage{microtype}

\usepackage{inconsolata}

\usepackage{graphicx}
\usepackage{amsmath}

\title{Where exactly does contextualization in a PLM happen?}

 \author{Soniya Vijayakumar, Simon Ostermann, Tanja Bäumel \and Josef van Genabith 
 \\
         German Research Instutite for Artificial Intelligence (DFKI), Saarland Informatics Campus, Germany \\
        {soniya.vijayakumar, simon.ostermann, tanja.baeumel, josef.van\_genabith
        }@dfki.de }

\begin{document}
\maketitle

\section{Introduction}
Pre-trained Language Models (PLMs) have shown to be consistently successful in a plethora of NLP tasks due to their ability to learn contextualized representations of words \cite{DBLP:journals/corr/abs-1909-00512}. BERT \cite{DBLP:journals/corr/abs-1810-04805}, ELMo \cite{peters-etal-2018-deep} and other PLMs encode word meaning via textual context, as opposed to static word embeddings, which encode all meanings of a word in a single vector representation. Much recent work has explored understanding the syntactic, semantic or contextualization information present in the hidden-states or attention-heads of the various PLMs by extracting contextualized representations either from the last output layer or concatenating the mean of representations across all layers 
\cite{10.1145/3397271.3401075, ravfogel-etal-2020-unsupervised, zhao2020quantifying, 
10.1145/3397271.3401075, wiedemann2019does, DBLP:journals/corr/abs-1909-00512, marecek-rosa-2019-balustrades, clark-etal-2019-bert}.
In this work, we present a study that aims to localize where exactly in a PLM word contextualization happens. In order to find the location of this word meaning \textit{transformation}, we investigate representations of polysemous words in the basic $BERT_{uncased}$ 12 layer architecture \cite{DBLP:journals/corr/abs-1810-04805}, a masked language model trained on an additional sentence adjacency objective, using qualitative and quantitative measures. 

Our main contribution is that unlike previous work, we do not restrict ourselves to the output layer(s) of such networks, but also investigate different \textit{sub-layers} within each BERT encoder layer. We also apply Dimensionality Reduction (DR) techniques on these latent sub-layers to better visualize our qualitative findings. 

First, we confirm findings of \citet{DBLP:journals/corr/abs-1909-00512} and other works, indicating that higher layers of BERT exhibit higher degrees of contextualization. 
Second, by investigating contextualization also in \textit{encoder sub-layers}, we find for the first time that representations in the self-attention sub-layers exhibit stronger and earlier signs of contextualization as compared to the  activation and output sub-layers.

\section{Methodology}
\paragraph{Experimental Setup:} We use the Contextualised Polysemy Word Sense v2 Dataset \cite{haber-poesio-2020-word} which contains custom samples of polysemous words in sentential contexts. We feed two sentences for each polysemous word to BERT and extract sub-layer vector representations for each encoder layer. For each word, we arrive at a set of vectors: \textit{Self-Attention (SA) sub-layer, Activation (Acts) sub-layer, and Output sub-layer}. We also extract the BERT static word embeddings (layer-0) for understanding semantic divergence of the sub-layers.

\paragraph{Measures of Contextuality:}

\textit{Sub-Layer Similarity.} Let \textit{w} be the polysemous word that appears in a pair of sentences $\{\textit{s1, s}2\}$ at index \textit{i,j} in its respective sentences, $\{x_{s1}$, $x_{s2}\}$ the sub-layer vector representations of the model \textit{m}. The sub-layer similarity of word \textit{w} in layer \textit{l} is:

\begin{equation}
SubLayerSim_{x}(w^l_{i,j}) = \frac{cos(x^l_{i_{s1}}, x^l_{j_{s2}})}{||x^l_{i_{s1}}|| ||x^l_{j_{s2}}||}
\end{equation}
where \textit{x} is Self-Attention sub-layer, Activation sub-layer and Output sub-layer or $l = \{0,1,...,11\}$. 

\textit{Static Word Embedding Similarity.}
For each word, we determine the cosine similarity between each sub-layer and its respective static word embedding from layer 0 (denoted as \textit{WESim}). 


\textit{Principal Components Analysis (PCA).} For better visualization, we reduce the High-Dimensional (HD) sub-layers (12 x 768, 12 x 3072) into two principal components using the PCA technique. PCA preserves the actual relative distance in the Euclidean data space and Principal Components (PCs) capture the direction of maximum variance. We determine squared L2 distances of the PCs between reduced sub-layer vector representations in the pair of sentences. We use these distance measures to quantitatively conform the observations made using \textit{SubLayerSim} similarities.

Intuitively, the higher the \textit{SubLayerSim} between the polysemous words, the closer the respective vector representations are to each other, hence they are less contextualized. The lower the \textit{WESim}, the further the vector is away from the (uncontextualized) static word embedding. On the other hand, the higher the \textit{L2 Distance}, the further the respective words (PCs) are from each other. This implies the actual relative distance in the higher dimension is high, indicating more contextualization.

\begin{figure}
\includegraphics[width=\linewidth]{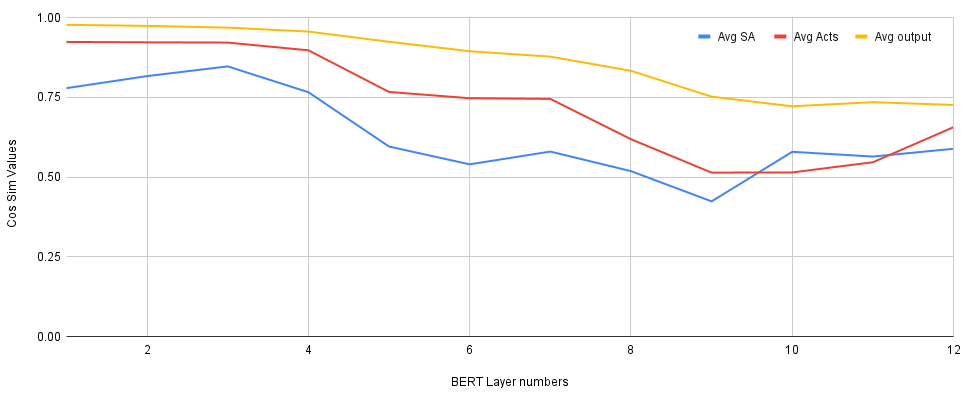}
\caption{Layer-wise average Self-Attention (SA), Activations (Acts) and Output \textit{SubLayerSim} for all words.} 
\label{fig:allwords}
\end{figure}

\begin{figure}
\includegraphics[width=\linewidth]{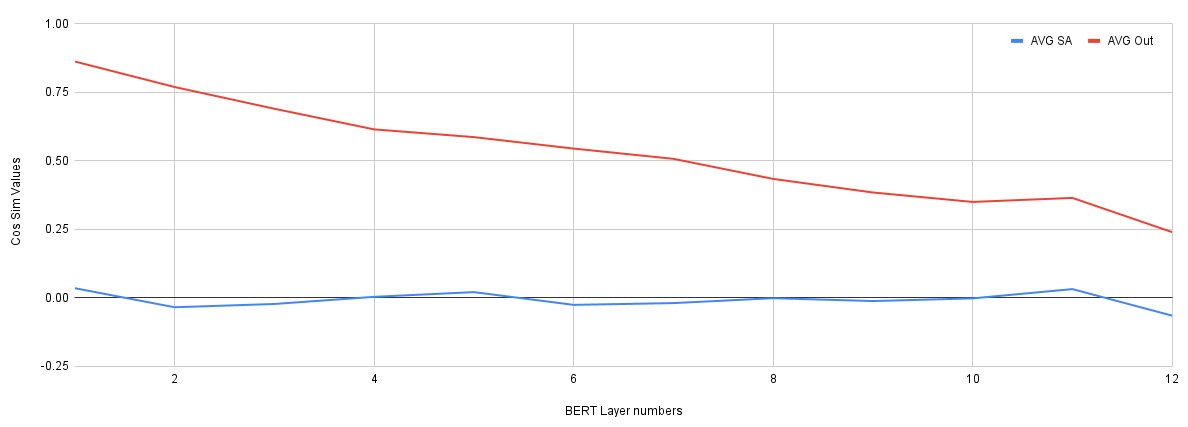}
\caption{Layer-wise average Self-Attention (SA) and Outputs \textit{WESim} for all words. the Acts sub-layer similarity is not reported due to the dimensionality mismatch.} 
\label{fig:weallwords}
\end{figure}

\begin{table}
\centering
\begin{tabular}{llll}
\hline
 & \textbf{Avg Sa} & \textbf{Avg Acts} & \textbf{Avg Outs} \\
\hline
\textbf{SLSim} & 0.6329 & 0.7309 & 0.8614\\
\textbf{WESim} & 0.008 & - & 0.521 \\
\textbf{L2 Dists} & 3.217 & 3.413 & 1.195\\
\hline
\end{tabular}
\caption{SLSim: SubLayerSim, WESim Word Embeddings and L2 distances for each sub-layer and all words.}
\label{tab:allwordsavg}
\end{table}

\section{Results \& Discussion}

Examining the average \textit{SubLayerSim} of all polysemous words, we observe that the Output sub-layer similarity is closer to one, indicating a lower degree of contextualization as compared to Acts and SA sub-layers (Figure \ref{fig:allwords}). The Acts and SA sub-layers \textit{SubLayerSim} are closer to each other, with higher contextualization in the SA sub-layer. The all-words average \textit{SubLayerSim} for the sub-layers indicate similar observations (Table \ref{tab:allwordsavg}). 

BERT's layer-wise average \textit{SubLayerSim} of all sub-layers decreases towards higher layers (lowest at layer 9), indicating the existence of higher contextualization layers between layers 5 and 9 (Figure \ref{fig:allwords}). This observation is similar to \citet{DBLP:journals/corr/abs-1909-00512}'s experiments: The upper hidden layers of contextualizing models produce more context-specific representations. An interesting observation is that the sub-layer similarities become higher and closer to each other in the penultimate layer of BERT (layer 12), which could be an indication of the model's sensitivity to output form and training objective. 
A notable finding in WESim is that, SA sub-layer WESim remains relatively consistent across BERT Layers where as the Output sub-layer WESim consistently decreases (Figure \ref{fig:weallwords}). This indicates the influence of residual connections in the BERT encoder Output sub-layer.




\paragraph{PCA:} We observe that the PCA bi-plots for SA, Acts and Output sub-layers are structurally different, indicating different structural alignment in their respective HD spaces. Examining the per-word average L2 distances, we observe that the Output sub-layer L2 distances are much lower than the respective SA and Acts sub-layers, indicating a stronger contextualization in the SA and Acts sub-layers (Table \ref{tab:allwordsavg}). 

\section{Conclusion \& Future (on-going) research}
In our research, we dive deeper into \textit{sub-layers} of each BERT encoder layer to localize the contextualization of polysemous words. Using contextuality measures and PCA, we observe that the contexts of these words are strongly captured in the Self-Attention sub-layer compared to the respective Activation and Output sub-layers and in higher layers of BERT. 
In other words, we find word meaning \textit{transformation} to occur strongly in SA and Acts sub-layer as compared to the Output sub-layers. We are currently working on exploring these methods on benchmark WSD corpora \cite{raganato-etal-2017-word} on other contextualized models. Our future research involves applying Representational Similarity Analysis (RSA) across the sub-layers \cite{abnar-etal-2019-blackbox} to understand if contextualization redundancies exist in these sub-layers. 

\section*{Limitations}
In this work, we are faced with a few limitations. First, the available WSD benchmark datasets are not in the format we require and hence, we conducted our experiments with a relatively smaller dataset, as presented above. Second, with the limited dataset, all the sentences have a standard structure for each polysemous word, i.e, each polysemous word is in the second position after the stopword 'The'. We are yet to experiment with sentences that have different indices for polysemous words. Third, we use two different metrics, cosine similarity and euclidean distance on a PCA for understanding the extent of contextualization. These two metrics have different properties making them not directly comparable. \textit{SubLayerSim} measure helped us in quantifying the contextualization whereas we used PCA as a visualization technique, to qualitatively confirm our observations.

\bibliography{anthology,custom}

\begin{thebibliography}{12}
\expandafter\ifx\csname natexlab\endcsname\relax\def\natexlab#1{#1}\fi

\bibitem[{Abnar et~al.(2019)Abnar, Beinborn, Choenni, and
  Zuidema}]{abnar-etal-2019-blackbox}
Samira Abnar, Lisa Beinborn, Rochelle Choenni, and Willem Zuidema. 2019.
\newblock \href {https://doi.org/10.18653/v1/W19-4820} {Blackbox meets
  blackbox: Representational similarity {\&} stability analysis of neural
  language models and brains}.
\newblock In \emph{Proceedings of the 2019 ACL Workshop BlackboxNLP: Analyzing
  and Interpreting Neural Networks for NLP}, pages 191--203, Florence, Italy.
  Association for Computational Linguistics.

\bibitem[{Clark et~al.(2019)Clark, Khandelwal, Levy, and
  Manning}]{clark-etal-2019-bert}
Kevin Clark, Urvashi Khandelwal, Omer Levy, and Christopher~D. Manning. 2019.
\newblock \href {https://doi.org/10.18653/v1/W19-4828} {What does {BERT} look
  at? an analysis of {BERT}{'}s attention}.
\newblock In \emph{Proceedings of the 2019 ACL Workshop BlackboxNLP: Analyzing
  and Interpreting Neural Networks for NLP}, pages 276--286, Florence, Italy.
  Association for Computational Linguistics.

\bibitem[{Devlin et~al.(2018)Devlin, Chang, Lee, and
  Toutanova}]{DBLP:journals/corr/abs-1810-04805}
Jacob Devlin, Ming{-}Wei Chang, Kenton Lee, and Kristina Toutanova. 2018.
\newblock \href {http://arxiv.org/abs/1810.04805} {{BERT:} pre-training of deep
  bidirectional transformers for language understanding}.
\newblock \emph{CoRR}, abs/1810.04805.

\bibitem[{Ethayarajh(2019)}]{DBLP:journals/corr/abs-1909-00512}
Kawin Ethayarajh. 2019.
\newblock \href {http://arxiv.org/abs/1909.00512} {How contextual are
  contextualized word representations? comparing the geometry of bert, elmo,
  and {GPT-2} embeddings}.
\newblock \emph{CoRR}, abs/1909.00512.

\bibitem[{Haber and Poesio(2020)}]{haber-poesio-2020-word}
Janosch Haber and Massimo Poesio. 2020.
\newblock \href {https://aclanthology.org/2020.pam-1.17} {Word sense distance
  in human similarity judgements and contextualised word embeddings}.
\newblock In \emph{Proceedings of the Probability and Meaning Conference (PaM
  2020)}, pages 128--145, Gothenburg. Association for Computational
  Linguistics.

\bibitem[{Khattab and Zaharia(2020)}]{10.1145/3397271.3401075}
Omar Khattab and Matei Zaharia. 2020.
\newblock \href {https://doi.org/10.1145/3397271.3401075} {Colbert: Efficient
  and effective passage search via contextualized late interaction over bert}.
\newblock In \emph{Proceedings of the 43rd International ACM SIGIR Conference
  on Research and Development in Information Retrieval}, SIGIR '20, page
  39–48, New York, NY, USA. Association for Computing Machinery.

\bibitem[{Mare{\v{c}}ek and Rosa(2019)}]{marecek-rosa-2019-balustrades}
David Mare{\v{c}}ek and Rudolf Rosa. 2019.
\newblock \href {https://doi.org/10.18653/v1/W19-4827} {From balustrades to
  pierre vinken: Looking for syntax in transformer self-attentions}.
\newblock In \emph{Proceedings of the 2019 ACL Workshop BlackboxNLP: Analyzing
  and Interpreting Neural Networks for NLP}, pages 263--275, Florence, Italy.
  Association for Computational Linguistics.

\bibitem[{Peters et~al.(2018)Peters, Neumann, Iyyer, Gardner, Clark, Lee, and
  Zettlemoyer}]{peters-etal-2018-deep}
Matthew~E. Peters, Mark Neumann, Mohit Iyyer, Matt Gardner, Christopher Clark,
  Kenton Lee, and Luke Zettlemoyer. 2018.
\newblock \href {https://doi.org/10.18653/v1/N18-1202} {Deep contextualized
  word representations}.
\newblock In \emph{Proceedings of the 2018 Conference of the North {A}merican
  Chapter of the Association for Computational Linguistics: Human Language
  Technologies, Volume 1 (Long Papers)}, pages 2227--2237, New Orleans,
  Louisiana. Association for Computational Linguistics.

\bibitem[{Raganato et~al.(2017)Raganato, Camacho-Collados, and
  Navigli}]{raganato-etal-2017-word}
Alessandro Raganato, Jose Camacho-Collados, and Roberto Navigli. 2017.
\newblock \href {https://aclanthology.org/E17-1010} {Word sense disambiguation:
  A unified evaluation framework and empirical comparison}.
\newblock In \emph{Proceedings of the 15th Conference of the {E}uropean Chapter
  of the Association for Computational Linguistics: Volume 1, Long Papers},
  pages 99--110, Valencia, Spain. Association for Computational Linguistics.

\bibitem[{Ravfogel et~al.(2020)Ravfogel, Elazar, Goldberger, and
  Goldberg}]{ravfogel-etal-2020-unsupervised}
Shauli Ravfogel, Yanai Elazar, Jacob Goldberger, and Yoav Goldberg. 2020.
\newblock \href {https://doi.org/10.18653/v1/2020.blackboxnlp-1.9}
  {Unsupervised distillation of syntactic information from contextualized word
  representations}.
\newblock In \emph{Proceedings of the Third BlackboxNLP Workshop on Analyzing
  and Interpreting Neural Networks for NLP}, pages 91--106, Online. Association
  for Computational Linguistics.

\bibitem[{Wiedemann et~al.(2019)Wiedemann, Remus, Chawla, and
  Biemann}]{wiedemann2019does}
Gregor Wiedemann, Steffen Remus, Avi Chawla, and Chris Biemann. 2019.
\newblock \href {http://arxiv.org/abs/1909.10430} {Does bert make any sense?
  interpretable word sense disambiguation with contextualized embeddings}.

\bibitem[{Zhao et~al.(2020)Zhao, Dufter, Yaghoobzadeh, and
  Schütze}]{zhao2020quantifying}
Mengjie Zhao, Philipp Dufter, Yadollah Yaghoobzadeh, and Hinrich Schütze.
  2020.
\newblock \href {http://arxiv.org/abs/2004.12198} {Quantifying the
  contextualization of word representations with semantic class probing}.

\end{thebibliography}
\bibliographystyle{acl_natbib}

\appendix



\end{document}